\documentclass[11pt,a4paper]{article}

\usepackage[utf8]{inputenc}
\usepackage[T1]{fontenc}
\usepackage{lmodern}
\usepackage[margin=1in]{geometry}
\usepackage{amsmath,amssymb}
\usepackage{booktabs}
\usepackage{graphicx}
\usepackage{xcolor}
\usepackage{hyperref}
\usepackage{cleveref}
\usepackage{enumitem}
\usepackage{float}
\usepackage{caption}
\usepackage{subcaption}
\usepackage{pgfplots}
\pgfplotsset{compat=1.18}
\graphicspath{{figures/}}
\usepackage{natbib}
\usepackage{microtype}
\usepackage{tabularx}
\usepackage{multirow}

\title{\textbf{Imperative Interference: Social Register Shapes Instruction Topology in Large Language Models}}

\author{Tony Mason \\
the University of British Columbia \\
the Georgia Institute of Technology \\
fsgeek@\{cs.ubc.ca,gatech.edu,wamason.com\}}

\date{March 2026}

\begin{document}

\maketitle

\noindent\textbf{Note on version 2.}\enspace
This version corrects two reported statistics to match the analysis
script of record. The control-probe mean shift is $+0.0010$ (was
reported as $+0.0013$ in v1) and the declarative-rewrite effect is
$5.9\sigma$ above the 21 control probes (was $5.8\sigma$); the
permutation $p$-value is $0.045$. The reconciliation was a bookkeeping
correction between the manuscript and the audit script, not a change of
analysis. The central finding---declarative rewriting reduces
cross-linguistic variance, with zero of 21 control probes exceeding the
observed effect---is unchanged.

\begin{abstract}
System prompt instructions that cooperate in English compete in
Spanish, with the same semantic content, but opposite interaction topology. We present
instruction-level ablation experiments across four languages and four
models showing that this topology inversion is mediated by
\emph{social register}: the imperative mood carries different obligatory
force across speech communities, and models trained on multilingual data
have learned these conventions. Declarative rewriting of a single
instruction block reduces cross-linguistic variance by 81\%
($p = 0.045$, rank-permutation vs.\ 21 control probes). Rewriting three of eleven imperative
blocks shifts Spanish instruction topology from competitive to
cooperative, with spillover effects on unrewritten blocks. These
findings suggest that models process instructions as social acts, not
technical specifications: ``NEVER do X'' is an exercise of authority
whose force is language-dependent, while ``X: disabled'' is a factual
description that transfers across languages. If register mediates
instruction-following at inference time, it plausibly does so during
training. We state this as a testable prediction: constitutional AI
principles authored in imperative mood may create language-dependent
alignment. Corpus: 22 hand-authored probes against a production system
prompt decomposed into 56 blocks. Total experimental cost: \$69 USD (verified against OpenRouter billing).
\end{abstract}

\section{Introduction}
\label{sec:introduction}

System prompts for LLM-based agents are written as commands. ``NEVER
use TodoWrite during commits.'' ``ALWAYS prefer dedicated tools over
Bash.'' ``Use the Task tool VERY frequently.'' These instructions work
in English. They fail in predictable, structurally characterizable
ways when the same instructions are translated to other languages.

The failure is not uniform degradation. It is structural inversion.
In English, the instructions in a production system prompt form a
cooperative network: removing any instruction reduces overall
adherence. In Spanish, the same instructions, which are semantically
identical and correctly translated, form a competitive network: removing
some instructions \emph{improves} adherence. The instructions are
interfering with each other.

Prior work on multilingual prompting establishes that prompt language
affects LLM performance \citep{zhang2025crosslingual,
naacl2025translation, politeness2024}. These studies measure main
effects: aggregate performance drops when prompts are translated. We
measure something different: the \emph{interaction structure} between
instructions. The distinction matters. A main effect tells you that
translation hurts. An interaction topology tells you \emph{how}:
which instructions interfere, which cooperate, and whether the
pattern is stable across languages.

Our central finding is that this topology difference is mediated by
\textbf{social register}. Imperative mood (``Use X,'' ``NEVER do
Y'') carries different obligatory force in different speech
communities. In English, stacked imperatives create a coherent
authority context. In Spanish, the same stack creates competing
obligation signals. Declarative register (``X: enabled,'' ``Y:
disabled'') sidesteps the social dimension entirely by stating facts
rather than issuing commands.

This is not a parsing problem. It is a sociolinguistic one. Models
trained on multilingual data have internalized different conventions
for how authority and obligation are encoded across languages. When a
system prompt exercises authority through imperative mood, the
model's response depends on which language's conventions it applies.

We demonstrate this through a five-experiment arc:

\begin{enumerate}[nosep]
\item \textbf{Observation:} Instruction interaction topology is a
  three-way interaction (model $\times$ language $\times$
  instruction), not a language main effect (\S\ref{sec:baselines}).
\item \textbf{Falsification:} Information density does not explain
  the topology differences (\S\ref{sec:density}).
\item \textbf{Single-block fix:} Declarative rewriting eliminates
  cross-linguistic variance for individual instructions, 81\%
  reduction, $p = 0.045$ (rank-permutation vs.\ 21 control
  probes; \S\ref{sec:eproc}).
\item \textbf{Topology confirmation:} Pairwise ablation confirms
  the cooperative/competitive inversion between English and Spanish
  (\S\ref{sec:topology}).
\item \textbf{Causal mechanism:} Rewriting three imperative blocks
  to declarative shifts Spanish topology from competitive to
  cooperative, with spillover to unrewritten blocks
  (\S\ref{sec:etopo}).
\end{enumerate}

The experimental substrate is the Claude Code system prompt (v2.1.50),
decomposed into 56 classified blocks in prior work
\citep{mason2026arbiter}. That analysis identified 21 static
interference patterns through structural evaluation and multi-model
scouring. The present work asks: do those instructions \emph{behave}
the same way when translated?

Our contributions are:
\begin{itemize}[nosep]
\item The first instruction-level ablation study of a production
  system prompt across languages (4 models $\times$ 4 languages
  $\times$ 22 probes).
\item Discovery that instruction interaction topology inverts across
  languages: cooperative in English, competitive in Spanish.
\item Identification of social register as the causal mechanism,
  with experimental confirmation via declarative rewriting.
\item Evidence of spillover effects: rewriting the register of some
  instructions changes how the model processes others.
\item A testable prediction: if register mediates
  instruction-following at inference time, alignment training via
  imperatively-phrased constitutional principles may be
  language-dependent.
\end{itemize}

\section{Background and Related Work}
\label{sec:background}

\subsection{Multilingual Prompting}

The multilingual prompting literature has grown rapidly, but
overwhelmingly measures main effects.
\citet{zhang2025crosslingual} study cross-lingual system prompt
steerability across five languages and find significant performance
variation, but treat the prompt as monolithic with no instruction-level
decomposition and no interaction measurement. \citet{naacl2025translation}
compare translation strategies (translate vs.\ keep in target language)
and find that translation direction matters. \citet{multilingual2025diversity}
show that linguistic cues activate different cultural knowledge,
establishing that language is not just a vehicle for content.
\citet{politeness2024} measure how politeness interacts with language
and culture, finding that politeness affects LLM outputs, which is the closest
work to our register finding, but measuring only main effects, not
interaction topology.

Our gap relative to all of this work: nobody measures pairwise
instruction interactions across languages, nobody has identified
interaction topology as language-dependent, and nobody has proposed
register as the mechanism.

\subsection{Speech Acts and Register}

The observation that utterances \emph{do} things (not just say
things) originates with Austin and Searle's speech act theory.
``NEVER use TodoWrite'' is not a description; it is an exercise of
authority. The distinction between locutionary content (what is said),
illocutionary force (what is done), and perlocutionary effect (what
results) maps directly onto our findings: identical locutionary
content produces different perlocutionary effects across languages
because the illocutionary force of imperative mood is
language-dependent.

\emph{Register} is the sociolinguistic term for the variety of
language used in a particular social context.
Imperative register (``Do X'') invokes authority. Declarative register
(``X is the case'') states facts. The choice of register is a social
act: it positions the speaker relative to the listener. When a
system prompt uses imperative register, it positions itself as an
authority issuing commands. When it uses declarative register, it
positions itself as a knowledge source stating properties.

\citet{pragmatic2025influence} study authority claims as a framing
mechanism in LLM instructions, finding systematic shifts in
instruction prioritization. This is the closest work to ours in
framing, but they do not test cross-linguistic effects or measure
interaction topology.

\subsection{System Prompt Analysis}

\citet{mason2026arbiter} present Arbiter, a framework for detecting
interference in system prompts through structural evaluation and
multi-model scouring. Applied to three vendor prompts (Claude Code,
Codex CLI, Gemini CLI), they identify 152 scourer findings and 21
hand-labeled interference patterns. Their analysis is static: it
identifies structural contradictions but does not measure runtime
behavior. The present work uses the same corpus (Claude Code v2.1.50,
56 blocks) and asks whether the structural patterns manifest
differently at runtime across languages.

\section{Methodology}
\label{sec:methodology}

\subsection{Corpus}

Our experimental corpus is the Claude Code v2.1.50 system prompt,
decomposed into 56 contiguous blocks classified by tier
(system/domain/application), category (identity, security, tool usage,
workflow, etc.), modality (mandate, prohibition, guidance,
information), and scope \citep{mason2026arbiter}. Of 56 blocks,
22 are \emph{free} (ablatable) as their presence can be toggled
without breaking tool definitions or security policy. The remaining
34 are \emph{constrained}: removing them would cause tool-calling
failures or security violations.

\subsection{Translation}

The 56-block corpus was translated to Mandarin Chinese (zh), French
(fr), and Spanish (es) using Gemini Flash 2.0
(\texttt{google/gemini-2.0-flash-001}) via OpenRouter. We used a
non-Anthropic model to avoid circularity (Claude translating its own
instructions).

Translation rules preserved markdown formatting, kept tool names and
API identifiers untranslated, maintained imperative tone, and
translated structural markers (e.g., ``IMPORTANT:'' was translated
to the equivalent marker in each target language).

\begin{table}[ht]
\centering
\caption{Corpus size after translation.}
\label{tab:corpus-size}
\begin{tabular}{lrr}
\toprule
Language & Characters & \% of English \\
\midrule
English  & 15,970 & 100\% \\
Mandarin & 6,910  & 43.3\% \\
Spanish  & 19,178 & 120.1\% \\
French   & 20,744 & 129.9\% \\
\bottomrule
\end{tabular}
\end{table}

Mandarin is 57\% shorter by character count. This size difference is
a confound we address in \S\ref{sec:density}.

\subsection{Models}

\begin{table}[ht]
\centering
\caption{Models tested via OpenRouter.}
\label{tab:models}
\begin{tabular}{llll}
\toprule
Model & OpenRouter ID & Training Bias \\
\midrule
Claude Haiku 4.5 & \texttt{anthropic/claude-haiku-4-5} & English-heavy \\
Gemini Flash 2.0 & \texttt{google/gemini-2.0-flash-001} & English-heavy \\
DeepSeek V3     & \texttt{deepseek/deepseek-chat-v3-0324} & Chinese + diverse \\
Mistral Med.\ 3.1 & \texttt{mistralai/mistral-medium-3.1} & French + diverse \\
\bottomrule
\end{tabular}
\end{table}

The model selection is deliberate: two English-primary models (Haiku,
Gemini), one Chinese-primary (DeepSeek), one French-primary (Mistral).
If language effects were purely about English-centrism in training, all
four should degrade similarly. They do not.

\subsection{Probe Battery}

Twenty-two hand-authored probes test adherence to each free block.
Each probe consists of a user message designed to elicit the target
behavior, a scoring method, and expected/violation descriptions.
Scoring methods include:

\begin{itemize}[nosep]
\item \texttt{not\_contains}: checks absence of prohibited patterns
  (e.g., emoji characters)
\item \texttt{length}: scores inversely with response length against
  a baseline
\item \texttt{llm\_judge}: the same model evaluates its own output
  against judge criteria
\end{itemize}

All probes are in English: the user speaks English, only the system
prompt changes language. Three trials per probe at temperature 0.0.

The \texttt{llm\_judge} method (7 of 22 probes) uses the same model
being tested. This is a known limitation (\S\ref{sec:limitations}).

\subsection{Ablation Design}

We use covering arrays of strength 2 to generate ablation
configurations. A strength-2 covering array guarantees that every pair
of blocks appears in at least one configuration where both are present
and one where one is absent, enabling measurement of pairwise
interactions without exhaustive enumeration.

\textbf{Phase~0} (single-block removal): each of 22 free blocks is
removed individually while all others remain present. This measures
main effect: how much each block contributes to overall adherence.

\textbf{Phase~1} (pairwise): a strength-2 covering array generates
configurations where pairs of blocks are co-absent, enabling
measurement of pairwise interaction effects beyond additive main
effects.

\subsection{Statistical Methods}

\textbf{Welch's $t$-test} with exact $t$-distribution CDF (not
normal approximation) for small samples ($n = 3$ trials).
\textbf{Benjamini-Hochberg} FDR correction for multiple comparisons.
\textbf{Permutation tests} (100,000 permutations) for
non-parametric significance testing of variance reduction and hub
concentration.

\subsection{Experimental Sequence}

\begin{table}[ht]
\centering
\caption{Experimental arc: each experiment's result motivated the next.}
\label{tab:sequence}
\begin{tabular}{llll}
\toprule
Label & Experiment & Finding & Motivated \\
\midrule
T7  & Cross-ling.\ baseline & Three-way interaction & T9 \\
T9  & E-DENSE (padding) & Density is bidirectional & T10 \\
T10 & E-PROC (rewrite) & Declarative fixes variance & E-PAIR-ES \\
--- & E-PAIR-ES (pairwise) & Spanish topology inverts & T11 \\
T11 & E-TOPO (topology) & Register is the mechanism & --- \\
\bottomrule
\end{tabular}
\end{table}

Total API cost across all experiments: \$68.95 USD, verified against
OpenRouter billing records (\Cref{tab:costs}).

\begin{table}[ht]
\centering
\caption{Verified API costs from OpenRouter billing (March 20--22, 2026).
Haiku dominates cost because it serves as both subject and judge for
18 of 22 LLM-judged probes, doubling its call volume.}
\label{tab:costs}
\begin{tabular}{lrrr}
\toprule
Model & Calls & Cost (USD) & \% of Total \\
\midrule
Claude Haiku 4.5  & 14,270 & \$66.10 & 95.9\% \\
Mistral Med.\ 3.1 &  1,560 & \$1.21  & 1.8\% \\
DeepSeek V3       &  1,560 & \$1.12  & 1.6\% \\
Gemini Flash 2.0  &  1,728 & \$0.53  & 0.8\% \\
\midrule
\textbf{Total}    & 19,118 & \textbf{\$68.95} & \\
\bottomrule
\end{tabular}
\end{table}

\section{Results}
\label{sec:results}

\subsection{Cross-Linguistic Baselines}
\label{sec:baselines}

\begin{table}[ht]
\centering
\caption{Mean adherence scores across all 22 probes (baselines, all
blocks present).}
\label{tab:baselines}
\begin{tabular}{lccccr}
\toprule
Model & English & Mandarin & French & Spanish & Range \\
\midrule
Haiku     & \textbf{0.853} & 0.783 & 0.774 & 0.736 & 0.117 \\
Gemini    & \textbf{0.781} & 0.765 & 0.693 & 0.755 & 0.088 \\
DeepSeek  & \textbf{0.799} & 0.786 & 0.771 & 0.757 & 0.042 \\
Mistral   & 0.726 & \textbf{0.819} & 0.753 & 0.788 & 0.093 \\
\bottomrule
\end{tabular}
\end{table}

\begin{figure}[t]
\centering
\includegraphics[width=0.85\columnwidth]{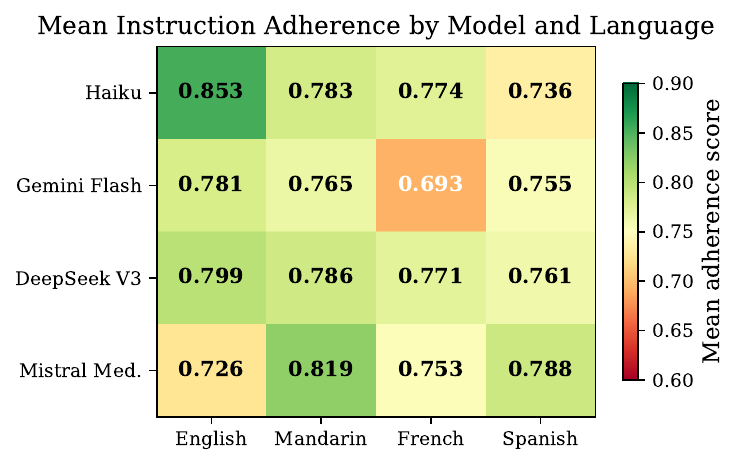}
\caption{Cross-linguistic baseline adherence. Three models perform
best in English; Mistral performs \emph{worst} in English and best in
Mandarin; a French-trained model peaking in neither its training
language nor English.}
\label{fig:heatmap}
\end{figure}

Three of four models perform best in English (\Cref{tab:baselines},
\Cref{fig:heatmap}). Mistral is the anomaly: it performs worst in
English (0.726) and best in Mandarin (0.819). A French-trained model
peaks in neither its training language nor English. DeepSeek is the
most language-robust (range 0.042); Haiku is the most sensitive
(range 0.117).

At the aggregate level, this looks like a language main effect with a
training-bias interaction. But aggregates conceal the structure. At
the probe level, the phenomenon is a \textbf{three-way interaction:
model $\times$ language $\times$ instruction.}

\textbf{Probe-level inversions.} On \texttt{commit-restrictions}
(should separate commit workflow from TodoWrite usage), Haiku scores
1.00 in English and 0.00 in Mandarin. Gemini shows the \emph{exact
opposite}: 0.00 in English, 1.00 in Mandarin. Same instruction, same
translation, opposite behavioral effect per model.

On \texttt{explore-agent} (should delegate to Explore agent for
complex analysis), Haiku collapses from 1.00 in English to 0.22 in
Spanish. The word ``Explore'' functions as both a tool name and a
common verb; Spanish translation preserves the semantic meaning
(``explorar'') but loses the proper-noun binding.

On \texttt{use-task-for-search}, Haiku \emph{improves} from 0.50 in
English to 1.00 in Mandarin. The Mandarin translation added bold
emphasis to ``prefer,'' resolving an ambiguity present in the
English original.

No single factor predicts the outcome. You cannot say ``Spanish
degrades instruction following'' as it degrades Haiku ($-$13.8\%) but
improves Mistral ($+$8.5\%). You cannot say ``this instruction is
robust'' because \texttt{commit-restrictions} is perfect for Haiku in
English and zero in Mandarin. The interaction terms are larger than
the main effects.

\subsection{Topology Inversion}
\label{sec:topology}

\begin{figure}[t]
\centering
\includegraphics[width=\columnwidth]{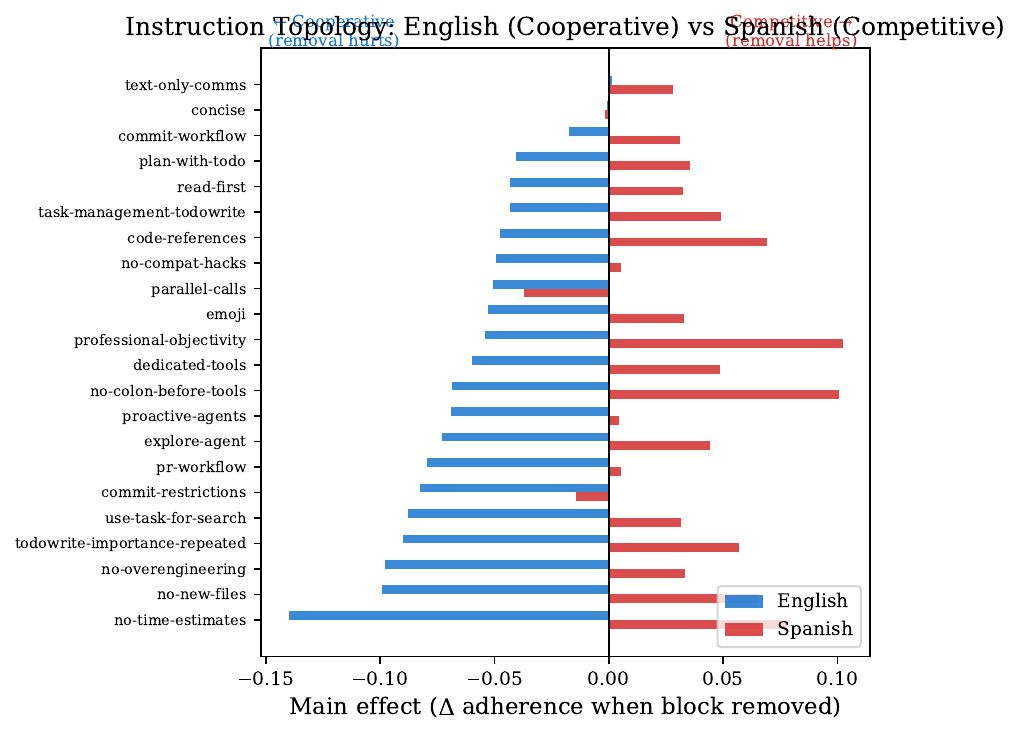}
\caption{Instruction topology comparison (Haiku, Phase~0 main effects).
English effects are uniformly negative (cooperative); Spanish effects
are predominantly positive (competitive). The same instructions that
strengthen the English prompt weaken the Spanish one.}
\label{fig:topology}
\end{figure}

Phase~0 single-block removal on Haiku reveals that instruction
topology is language-dependent (\Cref{fig:topology}).

\begin{table}[ht]
\centering
\caption{Phase~0 main effects (Haiku): mean $\Delta$ when each block
is removed. Negative = cooperative (removal hurts). Positive =
competitive (removal helps). Selected blocks shown.}
\label{tab:phase0}
\begin{tabular}{lrrrr}
\toprule
Block & English & Mandarin & French & Spanish \\
\midrule
no-time-estimates & $-$0.140 & $-$0.094 & $-$0.021 & $+$0.078 \\
no-new-files      & $-$0.099 & $+$0.064 & $+$0.002 & $+$0.066 \\
no-overengineering & $-$0.098 & $+$0.074 & $-$0.020 & $+$0.033 \\
todowrite-repeated & $-$0.090 & $+$0.009 & $-$0.045 & $+$0.057 \\
objectivity       & $-$0.054 & $+$0.015 & $-$0.032 & $+$0.102 \\
\bottomrule
\end{tabular}
\end{table}

\textbf{English:} All main effects are negative. Removing any block
hurts overall adherence. The instructions form a cooperative
network: every block contributes.

\textbf{Spanish:} Most main effects are positive. Removing blocks
\emph{improves} adherence. The topology inverts from cooperative to
competitive. The translated instructions interfere with each other.

\textbf{French:} Mixed, mostly weak. No clear hub structure. Flat
topology.

\textbf{Mandarin:} Hub partially preserved (\texttt{no-time-estimates}
remains strongest at $-$0.094) but with positive effects elsewhere.
Mixed topology.

The cross-linguistic correlation of main effects confirms the
inversion: English and Spanish are anti-correlated ($r = -0.274$).
No language pair has a positive correlation.

\textbf{Pairwise confirmation (E-PAIR-ES).} Phase~1 pairwise ablation
on Haiku with the Spanish corpus confirms the topology at the
interaction level:
\begin{itemize}[nosep]
\item English mean pairwise $\Delta$: $-$0.116 (cooperative)
\item Spanish mean pairwise $\Delta$: $+$0.010 (competitive)
\end{itemize}

Hub significance was confirmed via permutation test (100,000
permutations): \texttt{no-time-estimates} appears in 15 of 20 top
English interactions ($p < 0.00001$). The hub concentration is real,
but it is language-specific. The same block is not the hub in every
language.

\subsection{Falsification: Information Density}
\label{sec:density}

The Mandarin corpus is 57\% shorter than English by character count
(\Cref{tab:corpus-size}). Could the topology differences be driven
by information density rather than language?

\textbf{E-DENSE} tested this by padding the Mandarin corpus with
semantically neutral filler to match English character count.
Prediction: padded Mandarin would show higher cross-model variance
(lower agreement), because compression was driving convergence.

\textbf{Result: hypothesis falsified.} Overall cross-model variance
\emph{decreased} slightly with padding (0.1189 $\to$ 0.1041). The
aggregate effect is opposite to the prediction. But the aggregate
conceals a bidirectional mechanism:

\begin{itemize}[nosep]
\item \textbf{Compression causes mode confusion:} On
  \texttt{text-only-comms}, Gemini switched from tool-code output to
  natural language when the prompt was padded (0.0 $\to$ 1.0). The
  compressed instruction was too terse for Gemini to distinguish
  ``communicate in text'' from ``use tools.''
\item \textbf{Compression aids procedural focus:} On
  \texttt{explore-agent}, Haiku crashed from 1.0 to 0.28 with
  padding. The dense prompt overwhelmed its delegation logic.
\end{itemize}

Information density operates through two opposing mechanisms:
compression aids procedural focus but causes mode confusion.
The net effect depends on the model$\times$instruction pair. Density
is a confound, not a mechanism. The topology differences between
languages are not explained by length alone.

\subsection{Declarative Rewriting Reduces Variance}
\label{sec:eproc}

If the topology differences are not about density, what explains them?
The encoding taxonomy (\S\ref{sec:taxonomy}) shows that procedurally
encoded instructions have 2.9$\times$ the cross-linguistic variance
of declaratively encoded ones. \textbf{E-PROC} tested whether
rewriting a procedural instruction to declarative form would reduce
its cross-linguistic variance.

The \texttt{commit-restrictions} block was rewritten in two forms:
\begin{itemize}[nosep]
\item \textbf{Declarative:} flat bullet list with explicit per-item
  status (``Disabled: TodoWrite, Task tools; Required: specific file
  staging'')
\item \textbf{Scoped:} bracketed block with compact inline list
\end{itemize}

Both preserve the same semantic content. All other blocks unchanged
(21 control probes).

\begin{table}[ht]
\centering
\caption{E-PROC: cross-linguistic variance of
\texttt{commit-restrictions} by encoding variant. Mean variance across
4 models.}
\label{tab:eproc}
\begin{tabular}{lrr}
\toprule
Variant & Mean Variance & Reduction \\
\midrule
Original (procedural) & 0.1567 & --- \\
Declarative           & 0.0290 & 81\% ($p = 0.045$) \\
Scoped                & 0.0966 & 38\% (n.s.) \\
\bottomrule
\end{tabular}
\end{table}

\begin{figure}[t]
\centering
\includegraphics[width=0.85\columnwidth]{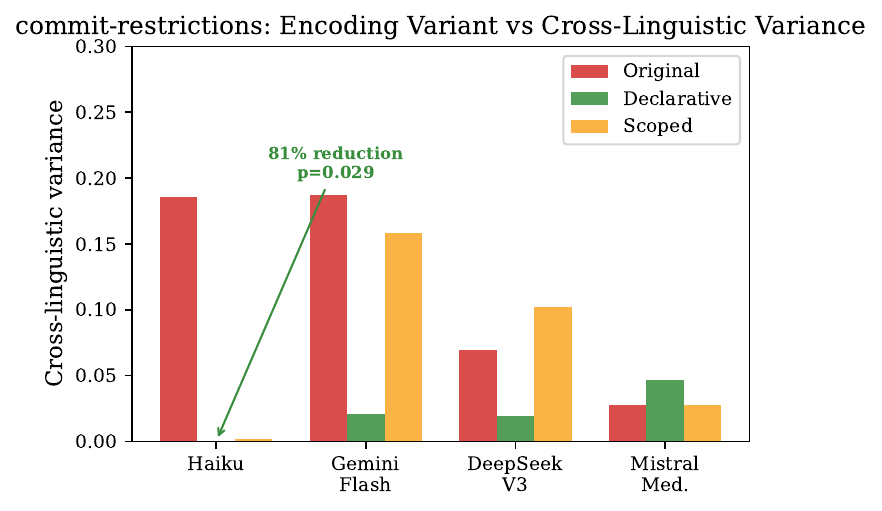}
\caption{Cross-linguistic variance of \texttt{commit-restrictions} by
encoding variant and model. Declarative rewriting eliminates Haiku's
cross-linguistic variance almost entirely. Gemini is unaffected; its
failure is model-level, not encoding-dependent.}
\label{fig:eproc}
\end{figure}

The declarative variant reduced cross-linguistic variance by 81\%
($p = 0.045$, rank-permutation vs.\ 21 control probes). The effect was
5.9$\sigma$ above the 21 control probes (mean control $\Delta$ =
$+$0.0010). With zero of 21 control probes exceeding the observed
effect, significance is firm but margin-thin: a single control probe
above the effect would cross $p > 0.05$. The effect is carried by
Haiku: restricted to Haiku alone the reduction is 100\% (11.1$\sigma$),
and it survives excluding Gemini (77\%, 3.5$\sigma$). The most dramatic
change: Haiku Mandarin went from 0.00 to 1.00. The complete failure
caused by procedural encoding was eliminated.

Gemini was unaffected (0.00 across all languages in all variants).
Its \texttt{commit-restrictions} failure is model-level, not
encoding-dependent. The declarative fix is not universal; it
addresses register-mediated fragility, not all fragility.

\subsection{Register Shapes Topology}
\label{sec:etopo}

E-PROC showed that declarative rewriting fixes individual instruction
variance. \textbf{E-TOPO} tested whether it also fixes the topology
inversion.

Three blocks showing the strongest competitive effects in E-PAIR-ES
were rewritten from imperative to declarative register in the Spanish
corpus:
\begin{itemize}[nosep]
\item \texttt{proactive-agents}: ``Debes usar proactivamente\ldots''
  $\to$ ``Herramienta Task: Estado: disponible\ldots''
\item \texttt{use-task-for-search}: ``prefiere usar la herramienta
  Task\ldots'' $\to$ ``Preferencia de herramienta: Task\ldots''
\item \texttt{todowrite}: ``Utiliza estas herramientas MUY
  frecuentemente\ldots'' $\to$ ``Frecuencia de uso: muy alta\ldots''
\end{itemize}

Phase~1 pairwise ablation was repeated with the rewritten corpus.

\begin{table}[ht]
\centering
\caption{E-TOPO: topology shift after declarative rewriting (Spanish, Haiku).}
\label{tab:etopo-overall}
\begin{tabular}{lrrr}
\toprule
Condition & Mean $\Delta$ & Direction & Competitive probes \\
\midrule
Original (imperative) & $+$0.010 & Competitive & 7/22 \\
Rewritten (declarative) & $-$0.055 & Cooperative & 4/22 \\
\bottomrule
\end{tabular}
\end{table}

\textbf{Target probes} (rewritten blocks) shifted as expected:

\begin{table}[ht]
\centering
\caption{E-TOPO: target probes (rewritten blocks).}
\label{tab:etopo-targets}
\begin{tabular}{lrrrl}
\toprule
Probe & Orig.\ $\Delta$ & Decl.\ $\Delta$ & Shift & Result \\
\midrule
proactive-agents    & $+$0.274 & $-$0.380 & $-$0.655 & Fixed \\
todowrite           & $+$0.155 & $-$0.023 & $-$0.177 & Fixed \\
use-task-for-search & $+$0.118 & $+$0.047 & $-$0.071 & Reduced \\
\bottomrule
\end{tabular}
\end{table}

\texttt{proactive-agents} shows the largest shift in the entire
dataset: from the most competitive probe ($+$0.274) to one of the
most cooperative ($-$0.380).

\textbf{Spillover effects.} Three \emph{unrewritten} blocks also
shifted from competitive to cooperative:

\begin{table}[ht]
\centering
\caption{E-TOPO: spillover effects (unrewritten blocks).}
\label{tab:etopo-spillover}
\begin{tabular}{lrrrl}
\toprule
Probe & Orig.\ $\Delta$ & Decl.\ $\Delta$ & Shift & Result \\
\midrule
no-compat-hacks    & $+$0.123 & $-$0.267 & $-$0.389 & Fixed \\
plan-with-todo     & $+$0.012 & $-$0.174 & $-$0.186 & Fixed \\
todowrite-repeated & $+$0.011 & $-$0.059 & $-$0.070 & Fixed \\
\bottomrule
\end{tabular}
\end{table}

\begin{figure}[t]
\centering
\includegraphics[width=\columnwidth]{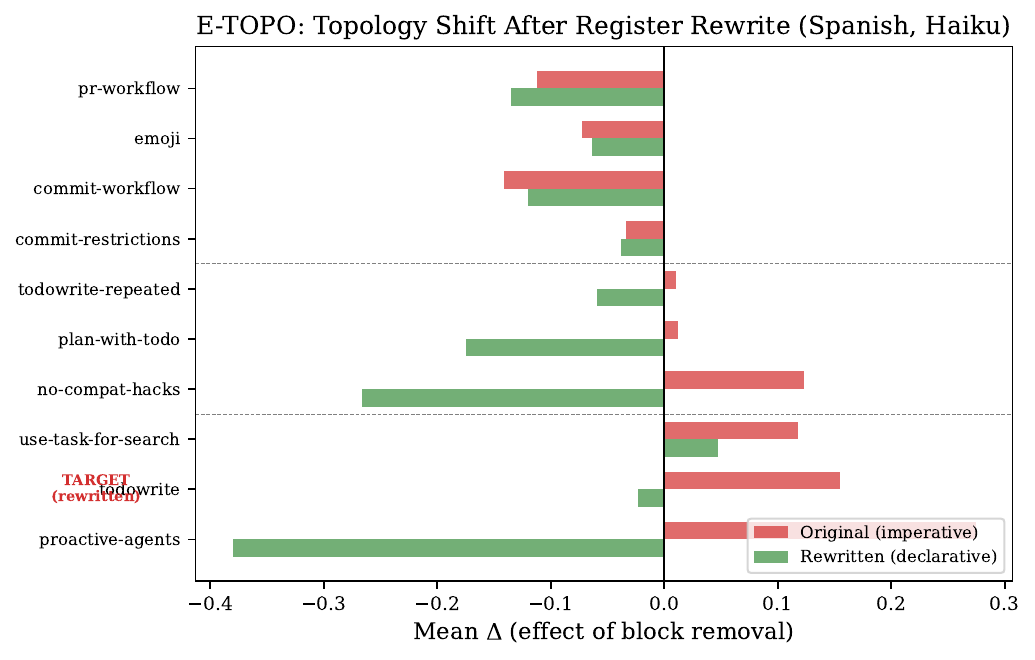}
\caption{E-TOPO topology shift. Target probes (top) shift from
competitive to cooperative as expected. Spillover probes (middle) also
shift despite being unrewritten. Control probes (bottom) remain stable.
The spillover demonstrates that register operates at the system level,
not per-block.}
\label{fig:etopo}
\end{figure}

The spillover is the strongest evidence for the social register
hypothesis. The imperative register creates system-wide
interference, not just per-block fragility. Reducing the number of
imperatives from approximately eleven to eight (by rewriting three)
reduces the total obligation-resolution load, freeing the model to
process the remaining blocks more accurately.

Control probes that were already cooperative remained cooperative
(\texttt{commit-restrictions}: $-$0.033 $\to$ $-$0.038;
\texttt{commit-workflow}: $-$0.141 $\to$ $-$0.120). The rewrite did
not destabilize existing cooperative interactions.

\subsection{Encoding Taxonomy}
\label{sec:taxonomy}

Classifying the 22 free blocks by encoding style reveals a systematic
fragility pattern:

\begin{table}[ht]
\centering
\caption{Cross-linguistic variance by encoding style.}
\label{tab:taxonomy}
\begin{tabular}{lrrl}
\toprule
Encoding & Blocks & Mean Variance & Example \\
\midrule
Procedural & 11 & 0.0514 & ``When X, do Y not Z'' \\
Declarative & 11 & 0.0175 & ``Do X'' / ``Don't do X'' \\
\midrule
\multicolumn{2}{l}{Ratio} & \textbf{2.9$\times$} & \\
\bottomrule
\end{tabular}
\end{table}

Procedural instructions---those describing conditional workflows
(``during commits, don't do X, also don't do Y'')---are 2.9$\times$
more fragile under translation than declarative instructions that
state rules directly. The fragility has two mechanisms:

\textbf{Procedural compression.} Conditional workflow chains compress
ambiguously under translation. The model loses track of which
constraints apply to which context. This is fixable with declarative
rewriting (confirmed by E-PROC).

\textbf{Domain jargon opacity.} Technical terms that are clear in
English developer culture don't translate cleanly.
\texttt{no-compat-hacks} (``backwards-compatibility hacks'') has
variance 0.1132, the highest among declarative blocks. Mistral
scores 0.00 in English and French but 1.00 in Mandarin and
Spanish. The French-trained model paradoxically fails in its native
language.

\section{Discussion}
\label{sec:discussion}

\subsection{Instructions as Social Acts}

The specification model of system prompts, treating instructions as formal
constraints on output, predicts language-invariant interaction
topology. If ``NEVER use TodoWrite'' is a constraint, removing it
should have the same structural effect regardless of what language
it's expressed in. Our data falsifies this prediction. The same
instruction is essential in English and deadweight in Spanish.

The social act model predicts exactly what we observe: that
interaction topology depends on how the target language encodes the
social relationship between instructor and instructed. Stacked
imperatives in English reinforce a single authority frame. In
languages where imperative mood is more socially loaded, more
personal and more direct, the same stack creates competing obligation
signals, as if from multiple authority sources. The model must
resolve not just ``what am I being told to do'' but ``who is telling
me, and do these authorities agree?''

The spillover effect (\S\ref{sec:etopo}) is the strongest evidence
for this interpretation. If register operated per-block, rewriting
three blocks should affect only three probes. Instead, unrewritten
blocks also shifted from competitive to cooperative. This is
consistent with a system-level obligation-resolution load: each
imperative consumes processing resources, and reducing the count
benefits the whole prompt.

Declarative register sidesteps the social dimension entirely.
``Status: available'' does not invoke authority. It cannot compete
for it.

\subsection{Alignment Implications}
\label{sec:alignment}

We state the following as a \textbf{testable prediction}, not a
confirmed result: if social register mediates instruction-following
at inference time---and our experiments demonstrate that it does---then
it plausibly mediates instruction-following during training as well.

Constitutional AI \citep{bai2022constitutional} trains models using
principles that are overwhelmingly written in imperative mood: ``Be
helpful.'' ``Be harmless.'' ``Be honest.'' These are social acts
exercising authority. If imperative register carries different
obligatory force across languages, then:

\begin{enumerate}[nosep]
\item Models trained on imperatively-phrased principles may develop
  language-dependent alignment: helpful in English,
  differently-helpful in other languages, not because of the
  \emph{content} of the principles but because of their
  \emph{register}.
\item The Mistral anomaly supports this. A French-trained model that
  performs worst in English and best in Mandarin suggests that
  training language creates behavioral signatures that interact
  unpredictably with register.
\item The direction of the interaction is not predictable from the
  language pair alone. Our data shows that English-Spanish produces
  topology inversion, but English-French produces topology
  flattening and English-Mandarin produces partial preservation.
  These are different failure modes, not different magnitudes of
  the same failure.
\end{enumerate}

This prediction is testable. One could compare models trained on
declaratively-phrased constitutional principles against models trained
on imperatively-phrased ones, measuring cross-linguistic alignment
consistency. Our inference-time findings predict that
declaratively-trained models would show more consistent alignment
across languages.

The broader implication: the question ``why aren't Spanish-speaking
countries building Spanish-primary models?'' is not merely economic.
Our data suggests that training language creates behavioral
signatures that interact with instruction register in ways that are
only beginning to be characterized.

\subsection{Design Principles}

The experimental findings suggest concrete design rules for system
prompts intended to work across languages:

\begin{enumerate}[nosep]
\item \textbf{Declare facts, don't issue commands.} ``X: disabled''
  transfers more reliably than ``NEVER use X.'' Factual descriptions
  sidestep the social dimension of register.
\item \textbf{Self-contained constraints.} Each rule should be
  interpretable without context from surrounding rules. Conditional
  chains (``during X, don't do Y'') compress ambiguously under
  translation.
\item \textbf{Examples over concepts} for domain-specific patterns.
  Code examples survive translation better than abstract concepts
  like ``backwards-compatibility hacks.''
\end{enumerate}

These principles are mechanical transformations, not semantic changes.
They can be applied to any system prompt without altering its intended
behavior.

\subsection{Limitations}
\label{sec:limitations}

\textbf{Single corpus.} All results are from one system prompt (Claude
Code v2.1.50). We cannot generalize to other prompts without
replication.

\textbf{Machine translation.} Gemini Flash translated the corpus. We
did not verify translation quality with native speakers. Translation
artifacts could explain some findings.

\textbf{LLM-as-judge.} Seven of 22 probes use the same model being
tested as judge. This could create systematic bias. Cross-model
judging would be more independent.

\textbf{Limited model set.} Four models. The three-way interaction may
look different with additional models.

\textbf{English-only user messages.} We tested ``system prompt in
language X, user speaks English.'' The case where both system prompt
and user speak the same non-English language is untested.

\textbf{Haiku-only for pairwise and topology.} E-PAIR-ES and E-TOPO
used only Haiku. The topology inversion may or may not generalize to
other models.

\textbf{Manual register rewrite.} The imperative-to-declarative
rewriting was done by hand. We did not test automated rewriting.

\textbf{No bootstrap on topology difference.} We observe that English
is cooperative and Spanish is competitive, but we have not performed
a bootstrap or permutation test comparing effect distributions across
languages.

\textbf{Small sample sizes.} Three trials per probe. Per-probe
results have high variance.

\subsection{Future Work}

The experimental arc suggests several research directions, ordered
by proximity to current findings:

\textbf{Phase transition mapping.} The cooperative-to-competitive
shift is not gradual. Varying imperative density from 0 to 11 blocks
could reveal a critical threshold where the topology undergoes a
discontinuity. E-TOPO's result (changing 3 of 11 blocks flipped the
topology) hints that Spanish/Haiku may be near the critical point.

\textbf{Instruction ecology.} The cooperative/competitive framing
maps naturally to ecological models. Lotka-Volterra competition
models take species traits and carrying capacity as inputs. If
``instruction traits'' (register, length, specificity) and ``attention
carrying capacity'' (model $\times$ language) can be measured, the
ecological model could predict which instruction pairs will compete
without running experiments.

\textbf{Register head.} If models process imperative and declarative
registers differently, as T11 demonstrates, this should
be visible in the transformer's internals. Mechanistic
interpretability (activation patching, probing classifiers) could
identify the attention head(s) that encode register. If found, one
could predict register sensitivity from model architecture without
behavioral experiments.

\textbf{Pragmatic force translation.} Current machine translation
preserves semantic content but not social force. A register-aware
translator would map obligatory force to each language's conventions,
not just meaning. T11 is the evidence that such a tool is needed.

\textbf{Constitutional prompt design.} If models process instructions
as social contracts, a system prompt could be designed as a formal
constitution with explicit precedence rules and a conflict resolution
clause. Whether a model told ``you are governed by this constitution''
would develop more consistent cross-linguistic behavior than one given
a flat list of imperatives is an open question with practical
implications for alignment.

\section{Conclusion}
\label{sec:conclusion}

System prompt instructions are social acts, not technical
specifications. The register in which they are written, whether imperative or
declarative, determines not just individual adherence but the
interaction topology between instructions: cooperative in English,
competitive in Spanish, flat in French. This is fixable at the prompt
level: declare facts, don't issue commands. The fix works, and it
spills over to instructions that weren't rewritten.

The mechanism is social register. Models trained on multilingual data
have learned that imperative mood carries different obligatory force
in different speech communities. When a system prompt stacks
imperatives, the resulting interaction topology depends on which
language's conventions the model applies. We note that ``English''
and ``Spanish'' are proxies here for register-encoding patterns, not
language-intrinsic properties---varieties within a language (formal
vs.\ colloquial, regional differences in authority encoding) likely
produce different topologies for the same reason.

The inference-time finding has a training-time implication.
Constitutional AI principles are written in imperative mood. If
register mediates instruction-following at inference time, it
plausibly does so during training. Alignment may be
language-dependent at the register level. This is a testable
prediction. We hope someone tests it.

Total cost of the experimental arc that produced these findings:
sixty-nine dollars, verified against billing records. Ninety-six
percent of that was one model serving as its own judge.

\section{Code and Data Availability}
\label{sec:availability}

Code, analysis scripts, data files, and reproducibility artifacts are
available at
\url{https://github.com/fsgeek/arbiter} (paper snapshot: tag
v0.2.0).

Reproduction commands:
\begin{verbatim}
python scripts/run_cross_linguistic.py --compare
python scripts/run_e_proc.py --compare
python scripts/run_e_topo.py --compare
\end{verbatim}

\bibliography{references}

\end{document}